\pdfoutput=1

\documentclass[11pt]{article}

\usepackage{acl}

\usepackage{times}
\usepackage{latexsym}

\usepackage[T1]{fontenc}

\usepackage[utf8]{inputenc}

\usepackage{microtype}

\usepackage{inconsolata}
\usepackage{stmaryrd}

\usepackage{algorithm}
\usepackage{algorithmic}
\usepackage{amsmath}
\usepackage{bbm}
\usepackage{multicol}
\usepackage{hyperref}
\usepackage[nameinlink]{cleveref}
\usepackage{subfigure}

\usepackage{amssymb}
\usepackage{pifont}

\usepackage{color}
\usepackage{tabularray}
\usepackage[table]{xcolor} 

\usepackage{newunicodechar}

\usepackage{caption}
\usepackage{subcaption}

\usepackage{blindtext}
\usepackage{graphicx}
\usepackage{makecell}

\usepackage{tabularx}
\usepackage{booktabs}

\usepackage{arydshln}

\usepackage{booktabs}
\usepackage[normalem]{ulem}
\useunder{\uline}{\ul}{}

\usepackage{lipsum}
\usepackage{tikz}
\usepackage{fancyvrb}
\usepackage{listings}
\usepackage{enumitem}
\usepackage{tcolorbox}
\usepackage{multicol}
\usepackage{siunitx}
\usepackage{xpatch}
\usepackage{arydshln} 
\usepackage{graphicx}    

\usepackage{multirow}  
\usepackage{pifont}
\usepackage{amsmath}

\usepackage{tikz}
\usetikzlibrary{shapes.geometric, arrows}

\tikzstyle{startstop} = [ellipse, minimum width=3cm, minimum height=1cm,text centered, draw=black]
\tikzstyle{process} = [rectangle, minimum width=3cm, minimum height=1cm, text centered, draw=black]
\tikzstyle{decision} = [parallelogram, minimum width=3cm, minimum height=1cm, text centered, draw=black]
\tikzstyle{arrow} = [thick,->,>=stealth]

\definecolor{hidden-red}{RGB}{205, 44, 36}
\definecolor{hidden-blue}{RGB}{194,232,247}
\definecolor{hidden-orange}{RGB}{243,202,120}
\definecolor{hidden-green}{RGB}{34,139,34}
\definecolor{hidden-pink}{RGB}{255,245,247}
\definecolor{hidden-black}{RGB}{20,68,106}
\definecolor{purple}{RGB}{144,153,196}
\definecolor{yellow}{RGB}{255,228,123}
\definecolor{hidden-yellow}{RGB}{255,248,203}
\definecolor{tkcolor}{RGB}{224,223,255}
\definecolor{darkblue}{rgb}{0, 0.40, 0.75}

\tcbset{
  aibox/.style={
    width=\linewidth,
    top=8pt,
    bottom=4pt,
    colback=blue!6!white,
    colframe=black,
    colbacktitle=black,
    center
  }
}
\newtcolorbox{AIbox}[2][]{aibox,title=#2,#1}

\title{Scaling Laws for Many-Shot In-Context Learning with Self-Generated Annotations}

 \author{Zhengyao Gu$^{{1}}$\thanks{Equal Contribution.}, Henry Peng Zou$^{1}\footnotemark[1]$, Yankai Chen$^{2}$, Aiwei Liu$^{3}$, Weizhi Zhang$^{1}$, Philip S. Yu$^{1}$
 \\  $^{1}$University of Illinois Chicago \quad $^{2}$Cornell University \quad $^{3}$Tsinghua University\\
 {\color{blue}\texttt{\{zgu24,pzou3,wzhan42,psyu\}@uic.edu}} \\
 { \color{blue}\texttt{yankaichen@acm.org} \quad \color{blue}\texttt{liuaw20@mails.tsinghua.edu.cn}} \\
}

\begin{document}
\maketitle
\begin{abstract}
The high cost of obtaining high-quality annotated data for in-context learning (ICL) has motivated the development of methods that use self-generated annotations in place of ground-truth labels. While these approaches have shown promising results in few-shot settings, they generally do not scale to many-shot scenarios. In this work, we study ICL with self-generated examples using a framework analogous to traditional semi-supervised learning, consisting of annotation generation, demonstration selection, and in-context inference. Within this framework, we propose a simple baseline that outperforms ground-truth ICL in zero-shot, few-shot, and many-shot settings. Notably, we observe a \textit{scaling law} with this baseline, where optimal performance is achieved with more than 1,000 demonstrations. To fully exploit the many-shot capabilities of semi-supervised ICL, we introduce IterPSD, an iterative annotation approach that integrates iterative refinement and curriculum pseudo-labeling techniques from semi-supervised learning, yielding up to 6.8\% additional gains on classification tasks. Code is available at: \textcolor{blue}{\url{https://anonymous.4open.science/r/semi-supervised-icl-FA07}}

\end{abstract}

\section{Introduction}\label{sec:intro}

\begin{figure}[t]
    \centering
    \includegraphics[width=\linewidth]{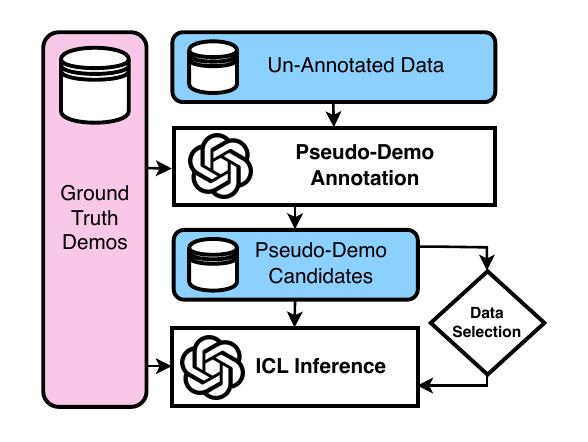}
    \caption{Semi-supervised ICL Framework. Ground truth data are used as demonstration for generating pseudo-demonstrations from unannotated data. The generated pseudo-demonstrations conjunctively  with a small ground truth demonstration, are selectively used as demonstrations for the final prompting.}
    \label{fig:flowchart}
\end{figure}

\begin{figure*}[ht]
    \centering
    \begin{subfigure}
        \centering
        \includegraphics[width=\linewidth]{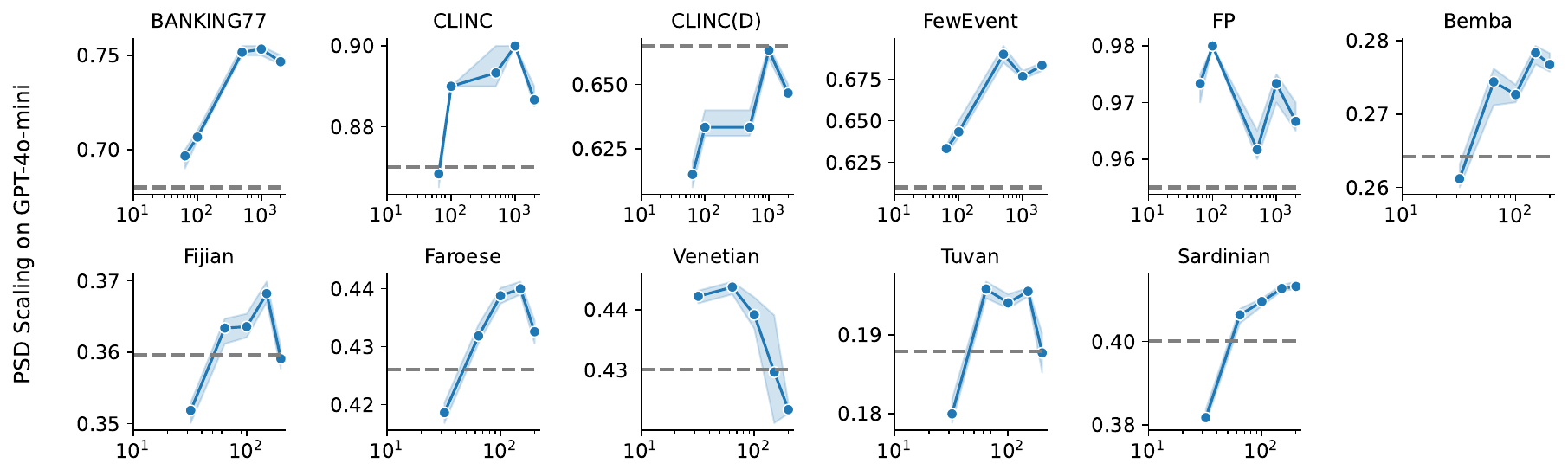}
    \end{subfigure}
    \hfill
    \begin{subfigure}
        \centering
        \includegraphics[width=\linewidth]{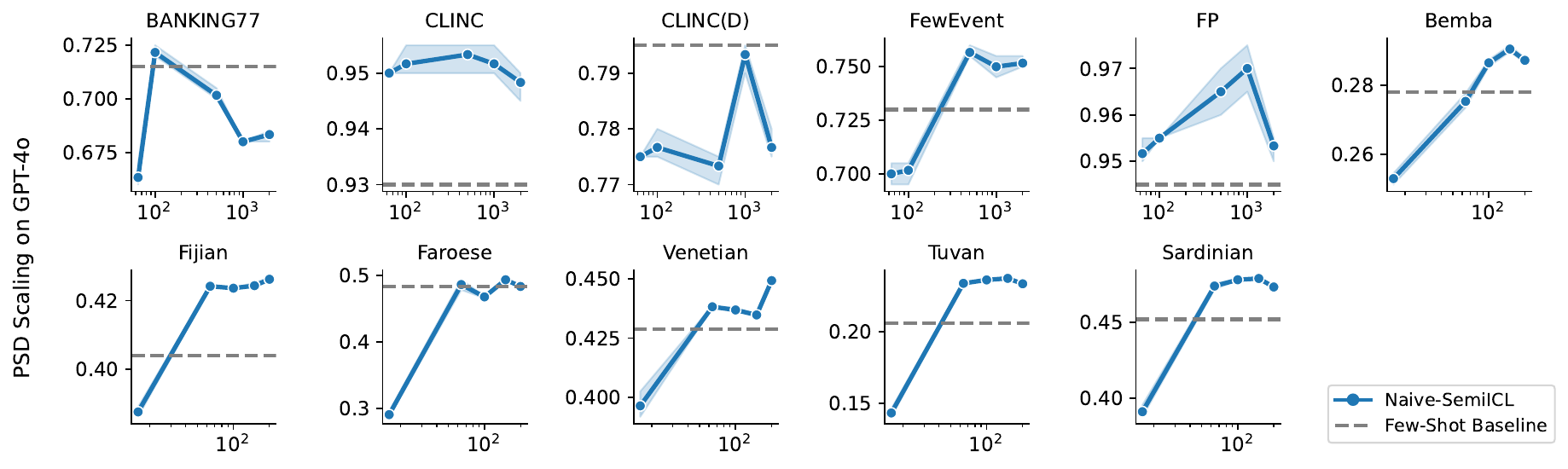}
    \end{subfigure}

    \caption{Scaling trend of Naive-SemiICL on classification and translation tasks with GPT-4o and GPT-4o-mini. The dashed gray line represents the few-shot baseline. Both model exhibits a scaling trend on most tasks. All experiments are performed with a ground truth budget of $k_l = 16$.}
    \label{fig:scaling}
\end{figure*}

In-context learning (ICL) has emerged as a powerful paradigm in natural language processing, enabling language models (LMs) to learn, adapt, and generalize from examples presented within their input context. This approach eliminates the need for extensive retraining and parameter modifications, facilitating more flexible and efficient learning \cite{icl-og, icl-work, many-shot-icl, fang2025tabgen}. The high cost of obtaining high-quality annotated data for ICL has motivated the development of methods \cite{auto-cot, li-qiu-2023-mot, pickle, self-aug-BLI, self-icl} that use self-generated annotations in place of ground-truth labels. However, previous research has not examined ICL performance with self-generated annotations in \textit{many-shot settings}. Recently, \cite{many-shot-icl} established a scaling law, showing that ICL performance improves with the number of demonstrations—up to thousands of examples. Inspired by this finding, we pose the following question:
\vspace{-0.5mm}
\begin{AIbox}{Research Question:}
    \textit{Can we scale ICL performance using self-generated demonstrations up to thousands of examples as well?}
\end{AIbox}
\vspace{-0.5mm}
\noindent We systematically investigate this question under a three-step framework (Figure~\ref{fig:flowchart}): \ding{172} annotation generation, \ding{173} demonstration selection, and \ding{174} semi-supervised inference, which we term \textit{Semi-Supervised ICL}. We first introduce a simple baseline, Naive-SemiICL, which annotates unlabeled data in a single iteration, scoring each annotation using the LLM’s verbalized confidence. Naive-SemiICL consistently outperforms ICL baselines in zero-shot, few-shot, and many-shot settings, as well as prior methods. We highlight that Naive-SemiICL achieves optimal performance with \textit{\textbf{1000 demonstrations}} on certain tasks (Figure~\ref{fig:scaling}).

With potentially thousands of self-annotated examples in the prompt, each demonstration can be viewed as a \textit{dataset}, which motivates the following question:
\vspace{-1mm}
\begin{AIbox}{Research Question:}
\textit{In what ways can techniques from traditional semi-supervised learning be leveraged to improve ICL performance?}
\end{AIbox}
\noindent We address this question by proposing \textit{IterPSD}, an iterative approach that progressively refines pseudo-demonstration quality by incorporating self-generated annotations at each iteration. IterPSD further improves semi-supervised ICL performance on five classification tasks, achieving gains of up to 6.8\% (Table~\ref{tab:iterpsd}).

\section{Method}\label{sec:method}

In this section, we establish the framework of Semi-Supervised ICL, which consists of three phases: \ding{172} pseudo-demonstration generation, \ding{173} demonstration selection, and \ding{174} semi-supervised inference. We then propose a simple baseline for Semi-Supervised ICL, Naive-SemiICL, which generates pseudo-demonstrations in a single iteration and filters out examples with low confidence scores. Building on Naive-SemiICL, we introduce an iterative method, IterPSD, that progressively improves the prompt by incorporating self-generated annotations during the demonstration generation process.

\subsection{Semi-Supervised ICL} 
\noindent \textbf{Confidence-Aware In-Context Learning} extends traditional ICL by outputting an additional confidence score for each input:
\begin{equation} \label{eq:confidence-aware-icl}
(y, r, c) = \mathrm{LM}(\rho_\mathcal{T}, \mathcal{E}, x)
\end{equation}
Like traditional ICL, the LLM is prompted with a task instruction $\rho_\mathcal{T}$ associated with task $\mathcal{T}$, a set of demonstrations $\mathcal{E}$, and an input $x$. Unlike traditional ICL, however, the model additionally returns a confidence score $c$ along with the predicted output $y$ and rationale $r$\footnote{In practice, generating rationales is optional. For example, one can query the LLM to directly generate the answer to a mathematical problem without intermediate reasoning steps.}, providing a measure of certainty for its predictions. We discuss the specific choice of confidence measure in Section \ref{sec:experiments}.\\

\noindent \textbf{Semi-Supervised ICL.} Beyond ground-truth annotations, Semi-Supervised ICL leverages unannotated data to enrich demonstrations. The setting assumes the availability of a \textit{ground-truth dataset} $\mathcal{D}_g = \{(x_i, y_i)\}^{N_l}$, usually small in quantity, alongside a large pool of \textit{unannotated data} $\mathcal{X}_u = {x_i}^{N_u}$. Semi-Supervised ICL augments the limited pool of ground-truth examples by generating \textit{pseudo-demonstrations} $\mathcal{D}_\mathrm{PSD}$ from the un-annotated data. Formally,
\begin{equation} \label{eq:psd_generation}
    \mathcal{D}_\mathrm{PSD} = \{(x, \Tilde{r}, \Tilde{y}, \Tilde{c}) \vert x \in \mathcal{X}_u\},
\end{equation}
where $(\Tilde{y}, \Tilde{r}, \Tilde{c}) = \mathrm{LM}(\rho_\mathcal{T}, \mathcal{E}_g, x)$ are generated by the LLM in an ICL fashion using a set of ground-truth demonstrated ns $\mathcal{E}_g \subseteq \mathcal{D}_g$. Low-confidence examples are filtered out according to a confidence threshold $\lambda$:
\begin{equation}\label{eq:psd_selection}
    \mathcal{D}_{\mathrm{PSD}}^\lambda = \{(x, \Tilde{r}, \Tilde{y}, \Tilde{c}) \vert \Tilde{c} \geq \lambda, (x, \Tilde{r}, \Tilde{y}, \Tilde{c}) \in \mathcal{D}_{\mathrm{PSL}}\}.
\end{equation}
During Semi-Supervised ICL inference, we sample pseudo-demonstrations $\mathcal{E}_\mathrm{PSD}$ from the filtered set $\mathcal{D}_{\mathrm{PSD}}^\lambda$, which we detail the specific methods in Appendix \ref{appendix:even_sampling}. The LLM is then prompted with these pseudo-demonstrations alongside the ground-truth demonstrations from which they were generated:
\begin{equation} \label{eq:inference}
    (\hat{y}, \hat{r}, \hat{c}) = \mathrm{LM}(\rho_\mathcal{T}, \mathcal{E}_l \cup \mathcal{D}_{\mathrm{PSD}}^\lambda, x).
\end{equation}
Most of the internal mechanisms of Semi-Supervised ICL are encapsulated by Equation \ref{eq:psd_generation}, where pseudo-demonstrations are generated, while Equation \ref{eq:psd_selection} and Equation \ref{eq:inference} represent simple operations. Next, we introduce two approaches for generating pseudo-demonstrations.

\subsection{A Simple Semi-Supervised ICL Baseline}

We propose a simple method that generates pseudo-demonstrations in a single iteration (Algorithm~\ref{alg:baseline}). We dub this method, along with the rest of the Semi-Supervised ICL framework, \textbf{Naive-SemiICL}. The method simply iterates over the unlabeled data for one iteration and generates a prediction, a rationale, and a confidence score for each input. As the simplest form of Semi-Supervised ICL, it provides effective in-context learning signals by filtering out low-quality pseudo-demonstrations. We experiment with three commonly used confidence measures on 16 datasets spanning 9 tasks, and show that this simple baseline consistently outperforms a strong 16-shot ICL baseline (Section~\ref{sec:Naive-SemiICL}).

\begin{algorithm}[!t]
\caption{Naive-SemiICL.}
\label{alg:baseline}
\begin{algorithmic}[1]

\STATE \textbf{Input:} prompt $\rho_\mathcal{T}$, ground-truth demonstrations $\mathcal{E}_l \subseteq \mathcal{D}_l$,
confidence score $\mathcal{C}$;

\STATE Initialize $\mathcal{D}_\mathrm{PSD} = \emptyset$;

\FOR{$x \in \mathcal{X}_u$}
    \STATE $\Tilde{y}, \Tilde{r}, \Tilde{c} = \mathrm{LM}(\rho_\mathcal{T}, \mathcal{E}_l, x)$;
    \STATE $\mathcal{D}_\mathrm{PSD} = \mathcal{D}_\mathrm{PSD} \cup \{(x, \Tilde{r}, \Tilde{y}, \Tilde{c})\}$;
\ENDFOR

\STATE \textbf{Return} $\mathcal{D}_\mathrm{PSD}$;

\end{algorithmic}
\end{algorithm}

\subsection{Iterative Pseudo-Demonstration Generation}
Encouraged by the success of Naive-SemiICL, we explore whether pseudo-demonstrations can enhance the accuracy of subsequent pseudo-demonstration generation. We propose \textbf{IterPSD} (Algorithm~\ref{alg:iterpsd}), an iterative method for generating pseudo-demonstrations that:

\begin{enumerate}
\item recursively adds newly generated pseudo-demonstrations to its own prompt until reaching the maximum number of allowed demonstrations (Line~\ref{alg:cap_demo}), and
\item re-samples the most confident pseudo-demonstrations according to a confidence threshold $\lambda$ from all previously annotated instances once the demonstration size reaches its limit (Line~\ref{alg:iterative_augment}).
\end{enumerate}

In each iteration, IterPSD samples and annotates $K$ unlabeled examples before applying a filtering step. The generated pseudo-demonstrations are recursively accumulated and fed back into the LLM to generate additional pseudo-demonstrations (Line~\ref{alg:iterpsd_single_iter}). To mitigate performance degradation caused by long context lengths, we impose an upper limit $\kappa$ on the number of self-fed pseudo-demonstrations. Once this limit is reached, we resample the $\kappa$ most confident pseudo-demonstrations from the generated pseudo-demonstrations, ensuring that only high-quality examples are retained (Line~\ref{alg:resample}).
\begin{figure*}[!thb]
    \centering
    \begin{subfigure}
        \centering
        \includegraphics[width=\linewidth]{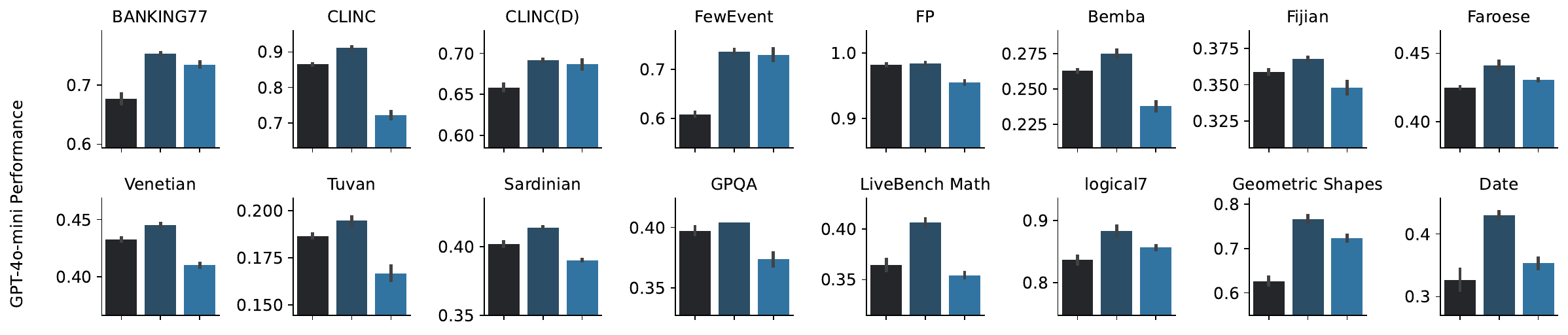}
    \end{subfigure}
    \hfill
    \begin{subfigure}
        \centering
        \includegraphics[width=\linewidth]{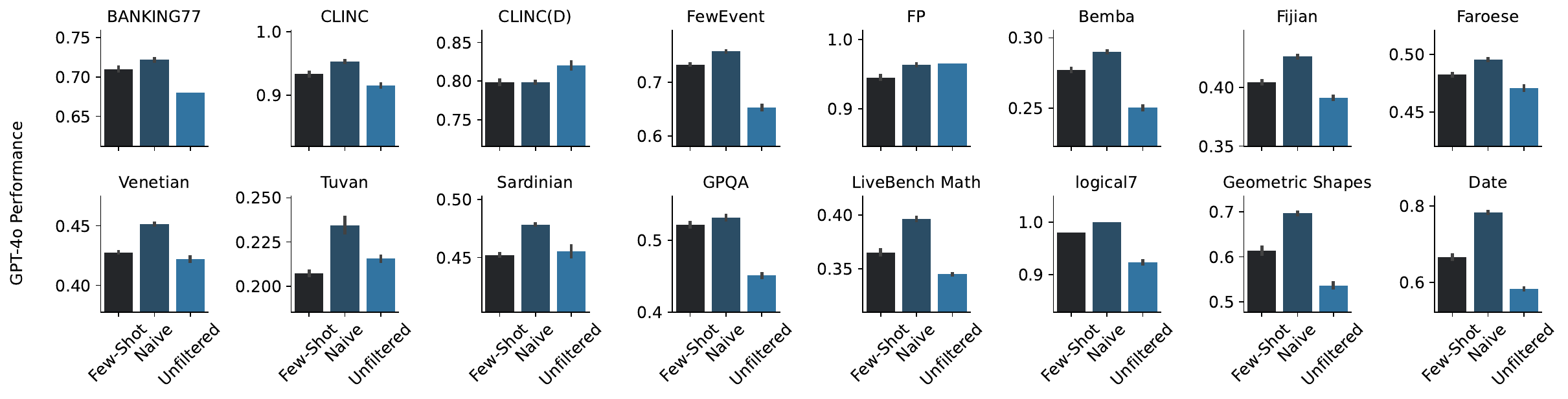}
    \end{subfigure}

    \caption{Comparison of GPT-4o-mini (top) and GPT-4o (bottom) performance across multiple datasets using three different methods: Few-Shot, Naive-SemiICL (Naive), and Naive-SemiICL without filtering (Unfiltered).}
    \label{fig:simple_baseline}
\end{figure*}

\paragraph{Order of Annotation.} To enhance the accuracy of pseudo-demonstration annotations, we introduce the $\epsilon$-Random Sampler (Algorithm \ref{alg:sampler}), a sampling strategy that selects both similar and diverse examples from the unannotated pool. At each iteration, a proportion $(1 - \epsilon)$ of examples is chosen based on their cosine similarity to the nearest previously annotated instances (Line \ref{alg:sample_sim})\footnote{We compute cosine similarity using vector embeddings generated by OpenAI's \textit{text-embedding-3-large}.}, ensuring that each selected example is similar to an existing annotation. The rest of the examples are chosen diversely in a clustering fashion similar to the one in \cite{auto-cot}. This approach that considers both the similarity and diversity aligns with curriculum learning \cite{curriculum-learning} in semi-supervised learning, which facilitate self-training by balancing a mixture of confident and uncertain predictions.

\paragraph{Mitigating Confirmation Bias.} To maintain annotation quality, we find that at least half of the data ($\epsilon \geq 0.5$) should be sampled diversely (Line \ref{alg:sample_random})\footnote{We found the best performing $\epsilon$ to be $0.8$ in most of our experiments.}. When $\epsilon = 0$, selections are exclusively based on similarity to previously annotated examples. The Pseudo-demonstrations become  homogeneous, leading to bias in ICL predictions. This phenomenon closely parallels confirmation bias in semi-supervised learning \cite{ssl-confirmation-bias, zou-caragea-2023-jointmatch}, highlighting a strong connection between Semi-Supervised ICL and traditional semi-supervised learning frameworks.
\begin{algorithm}[!t]
\caption{IterPSD}
\label{alg:iterpsd}
\begin{algorithmic}[1]

\STATE \textbf{Input:} prompt $\rho_\mathcal{T}$, ground-truth demonstrations $\mathcal{E}_l \subseteq \mathcal{D}_l$, chunk size $K$, ratio of random examples $\epsilon$, maximum number of pseudo-demonstrations $\kappa$,
confidence score $\mathcal{C}$;

\STATE Initialize $\mathcal{D}_\mathrm{PSD} = \emptyset$; {\color{gray} \COMMENT{\textit{\small Set of all the annotated pseudo-demonstrations.}}}
\STATE Initialize $\mathcal{D}_\mathrm{PSD}^\lambda = \emptyset$
\STATE Initialize $\overline{\mathcal{D}}_\mathrm{PSD} = \mathcal{X}_u$; {\color{gray} \COMMENT{\textit{\small Set of un-annotated data yet to be annotated.}}}

\WHILE{$\overline{\mathcal{D}}_\mathrm{PSD} \neq \emptyset$}
    \IF{$|\mathcal{D}_\mathrm{PSD}^\lambda| > \kappa$} \label{alg:cap_demo}
        \STATE $\mathcal{D}_\mathrm{PSD}^\lambda = $ top-$\kappa$ confident examples in $\mathcal{D}_\mathrm{PSD}$; \label{alg:resample}
    \ENDIF
    \\{\color{gray} \COMMENT{\textit{\small Cap the demonstration at a maximum size, prevent performance degradation from long-context.}}}
    \STATE $S = \mathrm{Sampler}(\mathcal{E}_\mathrm{Iter}, \overline{\mathcal{D}}_\mathrm{PSD}, K, \epsilon)$;
    \\{\color{gray} \COMMENT{\textit{\small Retrieves a sample of size $K$ using $\epsilon$-Random Sampler}}}
    \STATE $\mathcal{D}_S = \text{Naive-SemiICL}(S, \rho_\mathcal{T}, \mathcal{E}_l \cup \mathcal{D}_\mathrm{PSD}^\lambda)$; \label{alg:iterpsd_single_iter}
    \\{\color{gray} \COMMENT{\textit{\small One iteration of Naive-SemiICL.}}}
    \STATE $\mathcal{D}^\lambda_S = \{(x, \Tilde{r}, \Tilde{y}, \Tilde{c}) | (x, \Tilde{r}, \Tilde{y}, \Tilde{c}) \in \mathcal{D}_S, \Tilde{c} \geq \lambda\}$; {\color{gray} \COMMENT{\textit{\small Filter by confidence.}}}
    \STATE $\mathcal{D}_\mathrm{PSD}^\lambda = \mathcal{D}_\mathrm{PSD}^\lambda \cup \mathcal{D}^\lambda_S$; \label{alg:iterative_augment}
    \STATE $\mathcal{D}_\mathrm{PSD} = \mathcal{D}_\mathrm{PSD} \cup \mathcal{D}_S$;
    \STATE $\overline{\mathcal{D}}_\mathrm{PSD} = \overline{\mathcal{D}}_\mathrm{PSD} - \mathcal{D}_S$;
\ENDWHILE

\STATE \textbf{Return} $\mathcal{D}_\mathrm{PSD}$;

\end{algorithmic}
\end{algorithm}
\begin{algorithm}
\caption{$\epsilon$-Random Sampler}
\begin{algorithmic}[1]

\STATE \textbf{Input:} annotated demonstration $\mathcal{D}_l$, un-annotated demonstration $\overline{\mathcal{D}}_l$, chunk size $K$, random ratio $\epsilon$, prompt $\rho_\mathcal{T}$, embedder $\phi$.

\STATE Initialize $S = \emptyset$;
\STATE $K_\mathrm{random} = \epsilon K, K_\mathrm{sim} = (1 - \epsilon)K$;
\STATE Compute $d_{ij} = \mathrm{sim}_\mathrm{cos}(\phi(x_i), \phi(x_j))$ for all $x_i \in \mathcal{D}_l, x_j \in \overline{\mathcal{D}}_l$; 
\STATE Compute $d_j = \min_i d_{ij}$ for all $x_j \in \overline{\mathcal{D}}_l$; \\
{\color{gray} \COMMENT{\textit{\small Compute distance to the nearest annotated example.}}}
\STATE $S_\mathrm{sim} = \{x_j | d_j \in \text{Smallest}_{K_{\mathrm{sim}}} \{ d_j \} \}$; \label{alg:sample_sim}\\
{\color{gray} \COMMENT{\textit{\small select the $K_\mathrm{sim}$ examples with the smallest distance to its nearest annotated demonstrations}}} \label{alg:select_nearest_neighbor}
\STATE Compute $S_\mathrm{random}$, a random sample of size $K_\mathrm{random}$
from $\overline{\mathcal{D}}_l - S_\mathrm{sim})$; \label{alg:sample_random}
\STATE $S = S_\mathrm{sim} \cup S_\mathrm{random}$;
\STATE \textbf{Return} $S$;

\end{algorithmic}
\label{alg:sampler}
\end{algorithm}

\section{Experimetnal Setup} \label{sec:experiments}
\paragraph{Tasks and Datasets.} Our evaluation consists of 16 datasets spanning 9 tasks and 3 task types:
\begin{itemize}
    \item \textbf{Classification.} We include BANKING77 \cite{banking}, CLINC \cite{clinc}, FewEvent \cite{fewevent}, and FP \cite{fp}.
    \item \textbf{Translation.} We evaluate Naive-SemiICL's ability to translate English into low-resource languages using 6 datasets from FLORES200 \cite{flores200}: Bemba, Fijian, Faroese, Tuvan, Venetian, and Sardinian.
    \item \textbf{Reasoning.} We include 5 benchmarks spanning scientific, mathematical, and logical reasoning: GPQA \cite{gpqa}, LiveBench Math \cite{livebench}, and three tasks from BigBenchHard \cite{bbh}: Logical7, Geometric Shapes, and Date.
\end{itemize}
We describe these datasets in detail in Appendix~\ref{appendix:dataset_det} and explain how we split the training and testing data in Appendix~\ref{appendix:train-test-split}.

\noindent \textbf{Evaluation Metrics.} For all classification and reasoning tasks, we report \textbf{accuracy} as the performance metric. We evaluate the equivalence of LaTeX-style mathematical outputs on LiveBench Math using the parser described in \cite{eval-harness}. For translation tasks, we report the \textbf{ChrF++} score \cite{chrf} using its default configuration, as implemented in TorchMetrics \cite{torchmetrics}, following \cite{many-shot-icl}. We report the mean and standard error over three trials for baseline results \cite{gpt4o} (Section~\ref{sec:Naive-SemiICL}). The remaining results are based on a single trial.

\noindent \textbf{Baselines.} For baseline comparisons, we experiment with different pseudo-demonstration sizes for Naive-SemiICL: $k_u \in \{$32, 64, 100, 500, 1000, 2000$\}$ for classification tasks, and $k_u \in \{$32, 64, 100, 150, 200$\}$ for translation tasks.
\begin{itemize}
    \item \textbf{$k$-Shot ICL.} The LLM is prompted with $k$ ground truth annotated examples, where $k$ ranges from $0$ to $500$. Base on the number of ground truth annotation used, we divide our experiments into zero-shot (Table~\ref{tab:zero_shot})), few-shot (Fig.~\ref{fig:simple_baseline}), and many-shot (Fig.~\ref{fig:manyshot-ssl}) settings.
    
    \item \textbf{Unfiltered SemiICL.} To highlight the importance of confidence-based data selection, we include an unfiltered variant of Naive-SemiICL, which samples pseudo-annotations without applying the filtering step.
    
    \item \textbf{MoT.} \cite{li-qiu-2023-mot} We include MoT as a domain-specific baseline for reasoning tasks. Unlike MoT, Naive-SemiICL uses a simple one-step filtering mechanism for demonstration selection, whereas MoT requires querying the LLM for each example. Configuration details are provided in Appendix~\ref{appendix:baseline_det}.
    
    \item \textbf{Reinforced ICL.} \cite{many-shot-icl} demonstrate that prompting the LLM with self-generated reasoning chains filtered by ground-truth answers can significantly improve ICL performance. This method serves as an upper bound on semi-supervised ICL performance on the reasoning tasks when the filtering mechanism is assumed to be perfect.
\end{itemize}

\noindent During our preliminary experiments, we found Auto-CoT to be uncompetitive on our reasoning datasets, as it relies on simple heuristics for data selection that are no longer effective. Since MoT includes all of Auto-CoT’s steps except its entropy- and semantic-based filters, we opted not to include Auto-CoT in our experiments.


\paragraph{Confidence Scores.} We primarily evaluate three confidence metrics: Verbalized Confidence which prompts the LLM to generate the confidence score (Table~\ref{tab:prompts}), Entropy, and Self-Consistency. Self-Consistency measures the confidence as the frequency of the most frequent answer, and entropy is defined as
\begin{equation}
     c_{\mathrm{Ent}} = -\frac{1}{L} \sum_{i=s}^L \log P(w_i \mid w_{<i}).
\end{equation}



\begin{table}[!t]
\centering
\resizebox{\linewidth}{!}{%
\begin{tabular}{llccc}
\toprule
& \textbf{Task} & \textbf{Zero-shot} & \textbf{Naive} & \textbf{Improv.}\\
\midrule \midrule
\rule{0pt}{12pt} \multirow{5}{*}{Classification} & Banking & 61.50 & \textbf{78.00} & \color[HTML]{34a854}\textbf{26.8\%}\\
\rule{0pt}{12pt} & FewEvent & 56.00 & \textbf{65.00} & \color[HTML]{34a854}\textbf{16.07\%}\\
\rule{0pt}{12pt} & CLINC & 83.50 & \textbf{88.50} & \color[HTML]{34a854}\textbf{5.99\%} \\
\rule{0pt}{12pt} & CLINCD & 59.50 & \textbf{61.00} & \color[HTML]{34a854}\textbf{2.52\%} \\
\rule{0pt}{12pt} & FP & 91.00 & \textbf{94.50} & \color[HTML]{34a854}\textbf{3.85\%} \\
\midrule
\rule{0pt}{12pt} \multirow{6}{*}{Translation} & Bemba & 0.2437 & \textbf{0.2591} & \color[HTML]{34a854}\textbf{12.37\%} \\
\rule{0pt}{12pt} & Fijian & 33.63 & \textbf{35.16} & \color[HTML]{34a854}\textbf{4.55\%} \\
\rule{0pt}{12pt} & Faroese & 42.45 & \textbf{42.90} & \color[HTML]{34a854}\textbf{1.06\%} \\
\rule{0pt}{12pt} & Venetian & 42.03 & \textbf{42.82} & \color[HTML]{34a854}\textbf{1.88\%} \\
\rule{0pt}{12pt} & Tuvan & 16.17 & \textbf{18.40} & \color[HTML]{34a854}\textbf{13.79\%} \\
\rule{0pt}{12pt} & Sardinian & 37.06 & \textbf{38.46} & \color[HTML]{34a854}\textbf{3.78\%}\\
\midrule
\rule{0pt}{12pt} \multirow{5}{*}{Reasoning} & GPQA & 36.36 & \textbf{38.38} & \color[HTML]{34a854}\textbf{5.55\%} \\
\rule{0pt}{12pt} & Math & 35.48 & \textbf{36.58} & \color[HTML]{34a854}\textbf{3.10\%} \\
\rule{0pt}{12pt} & Logical7 & 65.00 & \textbf{72.00} & \color[HTML]{34a854}\textbf{10.77\%} \\
\rule{0pt}{12pt} & Shapes & 56.00 & \textbf{60.00} & \color[HTML]{34a854}\textbf{7.14\%} \\
\rule{0pt}{12pt} & Date & 40.00 & \textbf{65.00} & \color[HTML]{34a854}\textbf{62.5\%} \\
\bottomrule
\end{tabular}%
}
\caption{Performance comparison of Zero-shot ICL and Naive-SemiICL. All experiments are done on GPT-4o-mini. For Naive-SemiICL, we report the best performing number of pseudo-demonstrations.}
\label{tab:zero_shot}
\end{table}


\paragraph{Hyperparameters.} Unless stated otherwise, we filter all generated pseudo-demonstrations using the confidence threshold at the 90th percentile. We discuss the hyperparameters of IterPSD in Appendix~\ref{appendix:iterpsd_hyper}.

\paragraph{Models.} We experiment with GPT-4o-mini and GPT-4o, checkpointed on 2024-07-18 and 2024-11-20, respectively, for all of our experiments. We discuss the computational cost associated with our experiments in Appendix~\ref{appendix:comp_budget}.


\section{Empirical Analyses}
\subsection{Naive-SemiICL Consistently Beats Baselines} \label{sec:Naive-SemiICL}
We first compare the performance of Naive-SemiICL with Verbalized Confidence to the few-shot baseline. For Naive-SemiICL, we report performance using the optimal number of pseudo-demonstrations $k_u$ for each task. The best-performing $k_u$ values are shown in Tables~\ref{tab:best_shot} and~\ref{tab:best_shot_all}. Naive-SemiICL outperforms few-shot ICL on all tasks except CLINC(D), where it matches the baseline. Unfiltered SemiICL fails to match baseline performance in 20 out of 32 settings, highlighting the importance of the filtering step. A detailed breakdown of the best performance across different confidence scores is provided in Appendix~\ref{appendix:confidence}.
\begin{table}[!t]
\centering
\resizebox{\linewidth}{!}{%
\begin{tabular}{lccccc}
\toprule
\textbf{Method} & \textbf{GPQA} & \textbf{Math} & \textbf{Logical7} & \textbf{Shapes} & \textbf{Date} \\
\midrule \midrule
\rule{0pt}{12pt} Naive-SemiICL & \underline{42.42} & \textbf{40.78} & \textbf{90.00} & \textbf{78.00} & \textbf{79.00} \\
\rule{0pt}{12pt} MoT & \textbf{44.44} & 25.86 & 88.00 & \underline{64.00} & \underline{58.00} \\
\hline
\rule{0pt}{12pt} Reinforced ICL & 54.54 & 42.63 & 93.00 & 78.00 & 89.00 \\
\bottomrule
\end{tabular}%
}
\caption{Comparison of Naive-SemiICL (Naive) and
MoT on reasoning datasets using
GPT-4o-mini.}
\label{tab:reasoning}
\end{table}

We highlight the effectiveness of Naive-SemiICL in extremely low-resource settings through a zero-shot experimental design. We generate pseudo-demonstrations with \textit{no initial ground-truth demonstrations} and compare Naive-SemiICL to zero-shot prompting. The performance gap between Naive-SemiICL and the zero-shot baseline depends solely on the quality of the filtering mechanism. As shown in Table~\ref{tab:zero_shot}, Naive-SemiICL outperforms the zero-shot baseline on all tasks in the benchmark, attaining an average improvement of 11.36\% under GPT-4o-mini. This exceeds the average improvement of 9.94\% in the 16-shot setting (Figure~\ref{fig:simple_baseline}), suggesting that Naive-SemiICL is more effective resource-constrained conditions. 

Additionally, we found Naive-SemiICL to be effective in high-resource settings. Figure~\ref{fig:manyshot-ssl} compares the performance of Naive-SemiICL and ground-truth ICL when $k_l \in \{64, 100, 500\}$ ground-truth examples are available. Across three tasks, Naive-SemiICL consistently outperforms the corresponding $k$-shot baselines. We observe diminishing returns in performance gains as the number of annotated demonstrations increases. On average, $k_g = 64$ improves performance by 10.49\% over the baseline, whereas $k_g = 500$ yields only a 4.73\% improvement across the three tasks. Combining these results, Naive-SemiICL is most effective when ground-truth data is scarce, although it can still be effective in high-resource settings.

On reasoning datasets, Naive-SemiICL outperforms MoT on all tasks except GPQA, as shown in Table~\ref{tab:reasoning}. Surprisingly, the performance gap between the two methods is substantial on LiveBench Math, Shapes, and Date. We attribute this to two key differences between Naive-SemiICL and MoT: (1) MoT uses Entropy to filter low-quality demonstrations, which we show to be less reliable than Verbalized Confidence (see Table~\ref{tab:confidence}); and (2) in preliminary experiments, we found similarity-based retrieval to be less effective than diverse sampling. Naive-SemiICL samples diversely from a large pool of pseudo-demonstrations, which MoT is unable to do due to its requirement to query the LLM for each demonstration retrieval.

\subsection{Scaling Law for Semi-Supervised ICL} \label{sec:ssl-scaling}
 We observe a scaling law for Semi-Supervised ICL, similar to the one reported in many-shot ICL~\cite{many-shot-icl}, on classification and translation tasks. We illustrate this trend in Figure~\ref{fig:scaling}. Across all configurations, Naive-SemiICL performance improves with larger demonstration sizes, although the point of peak performance varies. Both GPT-4o and GPT-4o-mini scale effectively across most tasks, typically peaking between 500 and 1,000 examples for classification tasks and between 100 and 200 examples for translation tasks. GPT-4o exhibits a more stable scaling trend than GPT-4o-mini on translation tasks, with performance peaking later and declining more gradually. With more available ground-truth data, we also observe scaling trends on BANKING77 and FewEvent (Figure~\ref{fig:manyshot-ssl}).

\begin{table}[!t]
\centering
\resizebox{\linewidth}{!}{%
\begin{tabular}{lccccc}
\toprule
\textbf{Method} & \textbf{BANKING} & \textbf{CLINC} & \textbf{CLINC(D)} & \textbf{FewEvent} & \textbf{FP} \\
\midrule \midrule
\rule{0pt}{12pt}Naive-V & \underline{75.67} & 69.00 & 90.00 & 66.50 & 98.00 \\
\rule{0pt}{12pt}Naive-S  & 75.00 & \underline{73.50} & \underline{91.50} & 69.00 & \underline{98.00} \\
\rule{0pt}{12pt}\cellcolor{gray!18}Iter-V & \cellcolor{gray!18}\textbf{78.00} & \cellcolor{gray!18}69.00 & \cellcolor{gray!18}90.50 & \cellcolor{gray!18}\textbf{73.50} & \cellcolor{gray!18}98.00 \\
\rule{0pt}{12pt}\cellcolor{gray!18}Iter-S  & \cellcolor{gray!18}\textbf{78.00} & \cellcolor{gray!18}\textbf{78.50} & \cellcolor{gray!18}\textbf{94.50} & \cellcolor{gray!18}\underline{70.00} & \cellcolor{gray!18}\textbf{98.50} \\
\rule{0pt}{12pt}\textbf{Improvement} & \color[HTML]{34a854}\textbf{3.10\%} & \color[HTML]{34a854}\textbf{6.80\%} & \color[HTML]{34a854}\textbf{3.28\%} & \color[HTML]{34a854}\textbf{6.52\%} & \color[HTML]{34a854}\textbf{0.50\%} \\
\bottomrule
\end{tabular}%
}
\caption{Comparison of Naive-SemiICL (Naive) and IterPSD (Iter) methods on various datasets using GPT-4o-mini, evaluated using verbalized (-V) and self-consistency (-S) confidence scores. The best-performing results for each dataset are highlighted in bold, while the second-best results are underlined.}
\label{tab:iterpsd}
\end{table}

We hypothesize that Naive-SemiICL's decline in performance beyond a certain demonstration size stems from the accumulation of errors in pseudo-demonstrations. To isolate the negative impact of long contexts on the LLMs, we examine the scaling behavior when all demonstrations are ground-truth data. Figure~\ref{fig:manyshot-scaling} shows that both GPT-4o-mini and GPT-4o continue to improve as the number of demonstrations increases, even beyond the optimal demonstration size for Naive-SemiICL in the 16-shot setting. This suggests that the performance degradation is not caused by long context length, but rather by the accumulated errors in pseudo-demonstrations. This finding motivates the design of IterPSD, which addresses error accumulation in pseudo-annotations through curriculum learning and iterative refinement.

\begin{figure*}[!th]
    \centering
    \begin{subfigure}
        \centering
        \includegraphics[width=0.95\linewidth]{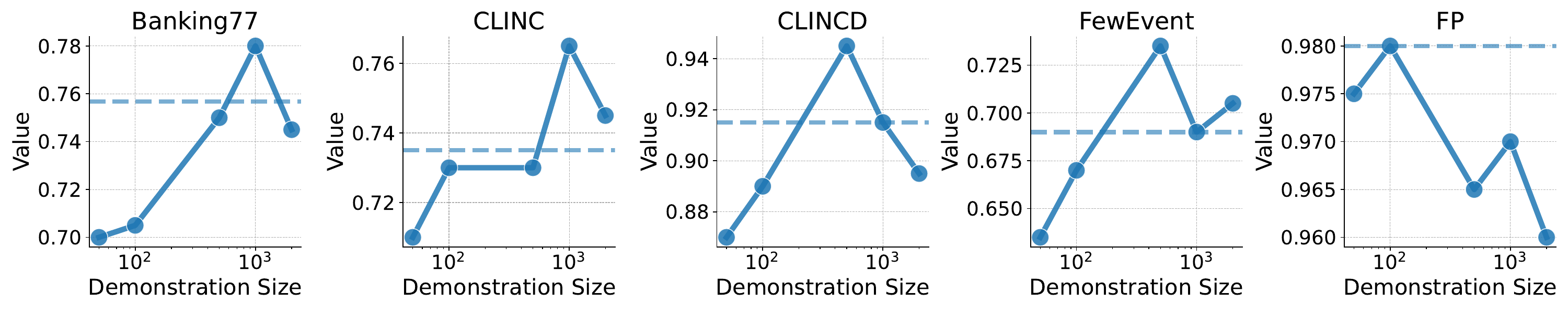}
    \end{subfigure}
    \caption{Scaling trend of IterPSD on five benchmark tasks. Blue horizontal dashed line represents the best performing Naive-SemiICL on the same dataset.}
    \label{fig:iterpsd_scaling}
\end{figure*}
\subsection{IterPSD Improves Upon Naive-SemiICL}
IterPSD outperforms Naive-SemiICL across five classification tasks, as shown in Table~\ref{tab:iterpsd}. We evaluate both methods using Verbalized Confidence and Self-Consistency. Notably, IterPSD achieves significant gains on BANKING, CLINC, CLINC(D), and FewEvent (over 3.0\% performance gain), but not on FP. Similar to Naive-SemiICL, we observe a scaling law with respect to the number of pseudo-demonstrations used in IterPSD. Clear scaling trends are observed in four out of five tasks, as shown in Figure~\ref{fig:iterpsd_scaling}. On these tasks, IterPSD attains peak performance with 500 to 1,000 pseudo-demonstrations. The lack of scaling on FP may be attributed to the relative ease of the dataset, as Naive-SemiICL already achieved 98\% accuracy on this task.

We also benchmark IterPSD on translation tasks, but the improvement over Naive-SemiICL is not consistent. We attribute this to the fact that each iteration of IterPSD needs to accumulate at least 100 demonstrations to avoid bias from sampling noise. However, Semi-Supervised ICL typically degrades after approximately 200 demonstrations, resulting in IterPSD terminating after 2 to 3 iterations.




\section{Related Work}\label{sec:related_work}
\paragraph{Self-Generated Demonstrations.} 
Large Language Models (LLMs) exhibit remarkable zero-shot capabilities, allowing them to perform tasks without task-specific fine-tuning or prior examples. Their zero-shot predictions have proven to be effective sources of demonstration for in-context learning \cite{zero-shot-learner, zou2025testnuc}.

Auto-CoT \cite{auto-cot} prompts the LLM with self-generated rationales on diversely sampled inputs. Rationales consisting of more than five reasoning steps are excluded from the demonstration to maintain the simplicity and accuracy of the demonstration. Such task-specific heuristic does not generalize to most recently published datasets such as LiveBench Math, as most of the generated rationales contain more than five steps. \cite{li-qiu-2023-mot} builds on top of Auto-CoT with extra an extra step of semantic filtering. At each example during inference, the LLM is prompted to choose the demonstration for itself after retrieving the semantically relevant demonstrations through an embedding model. Like Auto-CoT, Reinforced ICL \cite{many-shot-icl} generates rationales for reasoning problems and filters out those leading to incorrect answers. While this method requires ground truths, our filtering method do so with self-generated confidence score.

PICLe \cite{pickle} generates new demonstrations by annotating unlabeled examples and filtering out those with incorrect named entity types through self-verification prompting. Similarly, SAIL \cite{self-aug-BLI} employs an annotation strategy for the bilingual lexical induction task, discarding predictions that fail to translate back to the original input. Both methods rely on task-specific filtering and require additional LLM queries for self-verification or back-translation. In contrast, our Verbalized Confidence approach is task-agnostic and requires only a single prompt for pseudo-labeling, significantly reducing inference overhead. Z-ICL \cite{zicl} leverages the zero-shot generative capability of large language models to synthesize demonstrations for subsequent in-context learning inference. In contrast, our approach assumes access to abundant unlabeled data and a small set of ground-truth labels, using the LLM only for annotation rather than for input generation.

\paragraph{Many-Shot ICL.} \cite{many-shot-icl} observed a significant performance increase in a variety of generative and discriminative tasks, as well as a scaling law between the number of examples in the demonstration and ICL performance. Our method hinges on this ability as our proposed method, Naive-SemiICL, fits at least 64 examples in the prompt. We report a similar scaling law for Semi-Supervised ICL in this work.

\paragraph{Traditional Semi-Supervised Learning.} Semi-supervised learning seeks to reduce reliance on labeled data by leveraging abundant unlabeled data to enhance model performance \cite{lee2013pseudo, fixmatch, zou2025glean}. Self-training \cite{mclachlan1975iterative, xie2020self} iteratively refines the model by using its own predictions on unlabeled data for training. Pseudo-labeling \cite{lee2013pseudo, fixmatch, zou-etal-2023-decrisismb, zou2023semi} employs confidence-based filtering, retaining only high-confidence pseudo-labels to reduce error propagation and confirmation bias. JointMatch \cite{zou-caragea-2023-jointmatch} further alleviates error accumulation by using two independently initialized networks that teach each other through cross-labeling. 
Our work is the first to integrate confidence filtering and leverage both labeled and pseudo-labeled data in an in-context learning framework.

\section{Conclusion}
We introduced a semi-supervised ICL framework that enhances self-generated annotations through confidence-based data selection and iterative annotation. Our analysis of Naive-SemiICL with increasing amounts of ground-truth data reveals diminishing returns—while additional ground-truth annotations improve performance, the relative contribution of pseudo-demonstrations decreases. This suggests that semi-supervised ICL is particularly effective in low-resource settings, yet remains beneficial even when more ground-truth data is available. We further identify a scaling law in semi-supervised ICL, showing that models achieve optimal performance with over 1,000 pseudo-demonstrations. Our simple semi-supervised method, Naive-SemiICL, outperforms a strong 16-shot ICL baseline, achieving an average performance gain of 9.94\% across 16 datasets. We also propose IterPSD, an iterative refinement approach for pseudo-demonstrations, which yields up to 6.8\% additional gains on classification tasks. 

\clearpage

\section{Limitation}
While this work investigates the potential of semi-supervised ICL, several limitations remain. First, the reliance on SOTA LLMs for ICL introduces substantial computational overhead, posing challenges for researchers and practitioners with limited resources. Second, although we have shown that the incorporation of pseudo-demonstration generation strategies enhances and improves ICL performance on two models, the effectiveness of our proposed method might be sensitive to the choice of model. Future work can explore more advanced confidence calibration techniques for pseudo-demonstration selection, such as adaptive thresholding. Additionally, noise-aware in-context learning remains an under-explored domain that could potentially improve the robustness of Semi-Supervised ICL.

\bibliography{custom}

\clearpage
\appendix
\begin{table*}[!th]
\centering
\resizebox{\textwidth}{!}{%
\begin{tabular}{@{}llcccc|cccc@{}}
\toprule
& & \multicolumn{4}{c}{\textbf{GPT-4o-mini}} & \multicolumn{4}{c}{\textbf{GPT-4o}} \\
\cmidrule(lr){3-6} \cmidrule(lr){7-10}
\textbf{Task Type} & \textbf{Task} & \textbf{Verbalized} & \textbf{Self-Consistency} & \textbf{Entropy} & \textbf{Back-Translation} & \textbf{Verbalized} & \textbf{Self-Consistency} & \textbf{Entropy} & \textbf{Back-Translation} \\ \midrule \midrule
\textbf{Classification} & BANKING & $\mathbf{75.33 \pm 0.20}$ & $75.16 \pm 0.20$ & - & - & $72.17 \pm 0.20$ & $\mathbf{72.30 \pm 0.20}$ & - & - \\
& CLINC & $89.16 \pm 0.80$ & $\mathbf{91.17 \pm 0.40}$ & - & - & $95.50 \pm 0.70$ & $\mathbf{95.80 \pm 0.90}$ & - & - \\
& CLINCD & $66.33 \pm 0.50$ & $\mathbf{69.17 \pm 0.20}$ & - & - & $\mathbf{79.33 \pm 0.20}$ & $77.80 \pm 0.20$ & - & - \\
& FewEvent & $69.33 \pm 0.50$ & $\mathbf{73.33 \pm 0.20}$ & - & - & $76.17 \pm 0.50$ & $\mathbf{77.17 \pm 0.20}$ & - & - \\
& FP & $97.50 \pm 0.50$ & $\mathbf{97.83 \pm 0.20}$ & - & - & $96.50 \pm 0$ & $\mathbf{97.83 \pm 0.20}$ & - & - \\
& \textbf{AVG} & $79.53$ & $\mathbf{81.33}$ & - & - & $83.93$ & $\mathbf{84.18}$ & - & - \\
\midrule \midrule 
\textbf{Translation} & Bemba & $\mathbf{27.93 \pm 0.10}$ & - & $26.66 \pm 0.20$ & $27.42 \pm 0.30$ & $\mathbf{29.16 \pm 0.20}$ & - & $27.65 \pm 0.20$ & $28.34 \pm 0.20$ \\
& Fijian & $\mathbf{36.70 \pm 0.20}$ & - & $35.96 \pm 0.10$ & $36.14 \pm 0.10$ & $\mathbf{42.67 \pm 0.40}$ & - & $41.42 \pm 0.30$ & $41.98 \pm 0.40$ \\
& Faroese & $\mathbf{43.97 \pm 0.20}$ & - & $42.32 \pm 0.20$ & $43.95 \pm 0.20$ & $\mathbf{49.69 \pm 0.40}$ & - & $48.01 \pm 0.40$ & $48.93 \pm 0.30$ \\  
& Venetian & $\mathbf{44.41 \pm 0.20}$ & - & $43.84 \pm 0.10$ & $43.26 \pm 0.20$ & $\mathbf{45.05 \pm 0.30}$ & - & $44.53 \pm 0.50$ & $44.67 \pm 0.40$ \\  
& Tuvan & $\mathbf{19.61 \pm 0.30}$ & - & $19.53 \pm 0.10$ & $19.02 \pm 0.20$ & $\mathbf{23.75 \pm 0.30}$ & - & $23.01 \pm 0.30$ & $22.57 \pm 0.40$ \\  
& Sardinian & $\mathbf{41.27 \pm 0.20}$ & - & $40.53 \pm 0.10$ & $40.63 \pm 0.20$ & $\mathbf{47.94 \pm 0.20}$ & - & $46.82 \pm 0.10$ & $47.85 \pm 0.30$ \\
& \textbf{AVG} & $\mathbf{35.65}$ & - & $34.81$ & $35.07$ & $\mathbf{39.71}$ & - & $38.57$ & $39.06$ \\
\midrule \midrule
\textbf{Reasoning} & GPQA & $40.40 \pm 0.50$ & $\mathbf{42.42 \pm 0.50}$ & $41.41 \pm 0.50$ & - & $\mathbf{52.52 \pm 0.50}$ & $47.47 \pm 0.50$ & $52.52 \pm 0.50$ & -\\
& LB Math & $\mathbf{40.78 \pm 0.30}$ & $35.52 \pm 0.50$ & $35.48 \pm 0.30$ & - & $36.33 \pm 0.80$ & $\mathbf{39.78 \pm 0.30}$ & $30.10 \pm 0.30$ & - \\
& logical7 & $\mathbf{90.00} \pm 0.50$ & $84.00 \pm 0$ & $86.00 \pm 0.50$ &  - & $98.00 \pm 0.50$ & $\mathbf{100.00 \pm 0.50}$ & $100.00 \pm 0.50$ & - \\
& Geometric & $70.00 \pm 0$ & $66.00 \pm 0$ & $\mathbf{78.00 \pm 0.50}$ & - & $61.00 \pm 0$ & $67.00 \pm 0$ & $\mathbf{70.00 \pm 0.50}$ & - \\
& Date & $\mathbf{42.00 \pm 0.80}$ & $32.00 \pm 0$ & $35.00 \pm 0$ & - & $\mathbf{68.00 \pm 0.80}$ & $65.00 \pm 0$ & $67.00 \pm 0.50$ & - \\
& \textbf{AVG} & $\mathbf{56.64}$ & $51.99$ & $55.18$ & - & $63.17$ & $63.85$ & $\mathbf{63.92}$ & -
\end{tabular}%
}
\caption{Comparison of GPT-4o-mini and GPT-4o performance using different confidence scores. Each task is evaluated using different inference strategies: Verbalized, Self-Consistency, Entropy, and Back-Translation (where applicable). Reported values on  represent average accuracy and ChrF++ with standard deviations.}
\label{tab:confidence}
\end{table*}

\section{Prompts} \label{appendix:prompt}
The prompts are illustrated in Table \ref{tab:prompts}. \{CAPITAL LETTERS\} enclosed in curly brackets are variables that are substituted during inference.


\section{Experimental Details} \label{appendix:exp_det}
\begin{figure*}[ht]
    \centering
    \begin{subfigure}
        \centering
        \includegraphics[width=\linewidth]{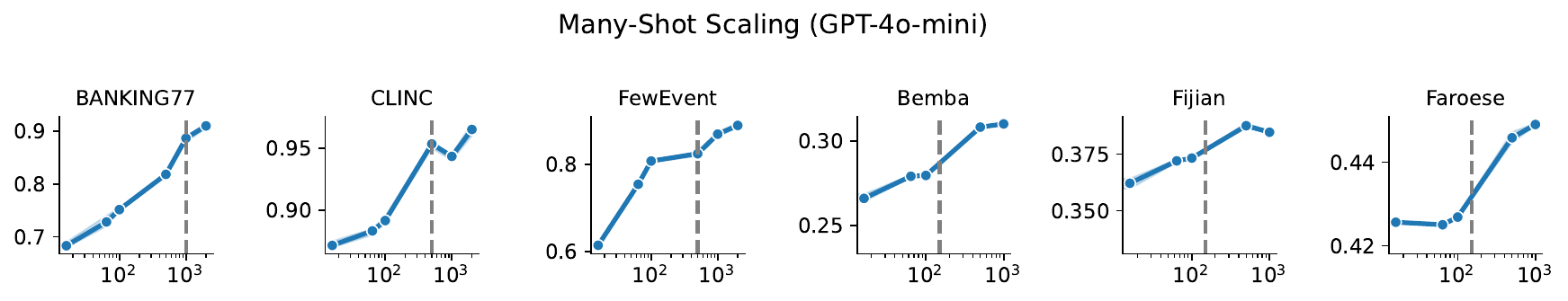}
    \end{subfigure}
    \hfill
    \begin{subfigure}
        \centering
        \includegraphics[width=\linewidth]{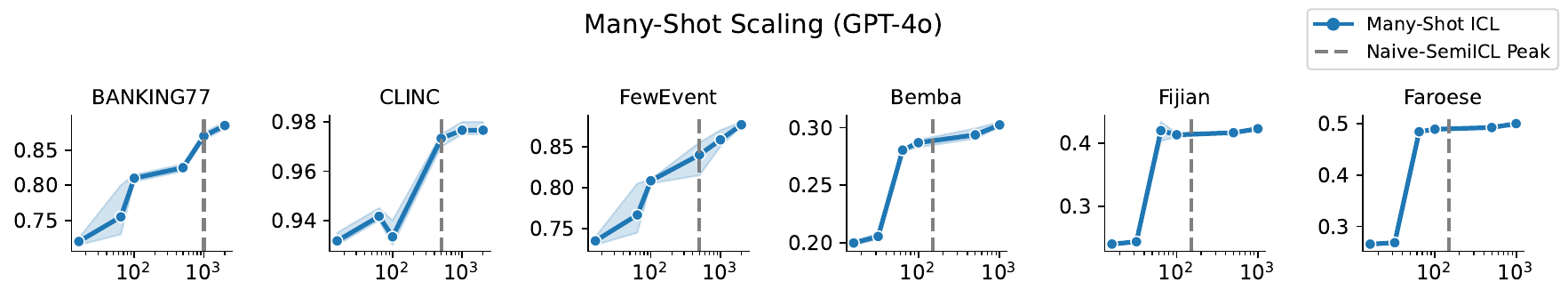}
    \end{subfigure}

    \caption{Many-shot scaling performance of GPT-4o-mini (top) and GPT-4o (bottom) across six selected datasets. The x-axis represents the number of shots (log scale), and the y-axis represents performance. The solid blue lines indicate many-shot in-context learning (ICL), while the dashed vertical lines mark the peak performance of Naive-SemiICL. Both models scale beyond the peak the performance of pseudo-demonstration approach.}
    \label{fig:manyshot-scaling}
\end{figure*}

\subsection{Train-Test Split} \label{appendix:train-test-split}
For classification tasks with more than 5,000 examples, we randomly sample 5,000 examples for demonstration and 200 for evaluation. For tasks with less than 5,000 examples, we randomly sample 200 for evaluation and use the rest for demonstration. Each FLORES dataset is comprised of a development set with 997 examples and a development test set with 1012 examples. We use all of 997 for demonstration and randomly sample 200 from the development test examples for evaluation. We use the diamond split (198 examples) of GPQA following \cite{many-shot-icl}, out of which 99 are used for evaluation and the other 99 are used for demonstration. Since LiveBench Math contains math problems from three sources, we evenly sample 150 questions from different sources for evaluation and use the rest for demonstration. Each BigBenchHard dataset contains 250 examples. We randomly sample 100 for evaluation and use the rest for prompting.\\

\subsection{Computational Budget} \label{appendix:comp_budget}
We ran all of our experiments on an Apple M3 chip, where embedding-based search constitutes less than 1\% of the computation time during IterPSD. The embeddings can be precomputed during data processing for each dataset, as it only needs to be computed once. It took about 400ms to retrieve each embeddings from the OpenAI API. The cost of generating the embeddings is \$0.13/million tokens. We ran all of our experiments on a \$1,000 budget.
\begin{figure*}[!th]
    \centering
    \includegraphics[width=\linewidth]{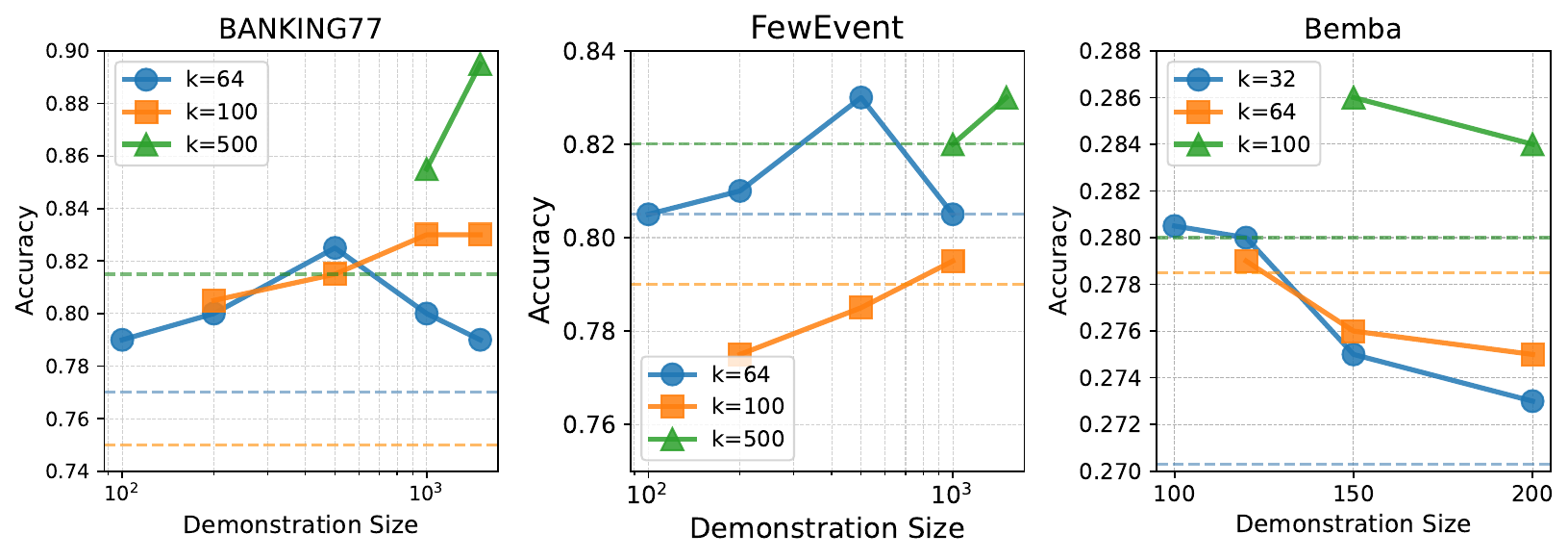}
    \caption{We compare Naive-SemiICL accuracy across different ground truth demonstration sizes, with baseline performances indicated by dashed lines. On FewEvent, the maximum number of pseudo-demonstrations is capped at 1000 due to the limited availability of pseudo-demonstrations after filtering.}
    \label{fig:manyshot-ssl}
\end{figure*}

\subsection{Dataset Details} \label{appendix:dataset_det}
\paragraph{Classification Datasets.}
\begin{itemize}
    \item \textbf{BANKING77.} The BANKING77\cite{banking} dataset is a fine-grained intent classification benchmark in the banking domain, consisting of 13,083 customer queries labeled into 77 intent categories. 
    \item \textbf{CLINC.} The CLINC150 \cite{clinc} dataset is a benchmark for intent classification, containing 22,500 user queries across 150 intent categories grouped into 10 domains, along with an out-of-scope category. We refer to the intent classification task of CLINC150 as CLINC.
    \item \textbf{CLINC(D).} We refer to the domain classification annotation of CLINC150 as CLINC(D).
    \item \textbf{FewEvent.} The FewEvent\cite{fewevent} dataset contains 4,436 event mentions across 100 event types, with each event type having only a few annotated examples (typically 5 to 10 per type).
    \item \textbf{FP.} Financial Phrasebank\cite{fp} The Financial PhraseBank dataset consists of 4840 sentences from English language financial news categorised by sentiment.
\end{itemize}

\paragraph{Low-Resource Language Translation.} FLORES-200 \cite{flores200} contains 200 languages translated from a common corpus. It is an extension of the original FLORES-101 \cite{flores101} dataset, which covered 101 languages. The dataset covers low-resource and high-resource languages, including many languages with little prior data on. It includes many African, South Asian, and Indigenous languages, making it one of the most diverse multilingual benchmarks. \\

\paragraph{Reasoning Datasets.}
\begin{itemize}
    \item \textbf{GPQA.} GPQA\cite{gpqa} is a multiple-choice question answering benchmark, with graduate-level questions that involves reasoning in biology, physics, and chemistry.
    \item \textbf{LiveBench Math.} LiveBenchMath contains 368 contamination-free mathematical problems, sampled from high school math competitions, proof-based fill-in-the-blank questions from Olympiad-level problems, and an enhanced version of the AMPS dataset.
    \item \textbf{BigBenchHard.} We include three tasks from BigBenchHard\cite{bbh}. \textbf{Logical7} evaluates a model's ability to deduce the order of a sequence of objects based on provided clues about their spatial relationships and placements. The \textbf{Geometric Shapes} task within the BigBenchHard evaluates a model's ability to interpret and identify geometric figures based on SVG path data. The \textbf{Date} task within the BigBenchHard benchmark evaluates a model's ability to comprehend and manipulate date-related information.
\end{itemize}

\begin{table}[!ht]
\centering
\resizebox{\linewidth}{!}{%
\begin{tabular}{lccccc}
\toprule
 & \textbf{GPQA} & \textbf{LiveBench Math} & \textbf{Logical7} & \textbf{Geometric Shapes} & \textbf{Date} \\
\midrule \midrule
GPT-4o-mini & 4 & 0 & 8 & 8 & 0\\
\hline
GPT-4o & 4 & 0 & 8 & 8 & 16 \\
\bottomrule
\end{tabular}%
}
\caption{Best number of shots for the baseline on reasoning tasks. We use the same number of shots for Naive-SemiICL.}
\label{tab:best_shot}
\end{table}

\begin{table*}
\centering
\resizebox{\linewidth}{!}{%
\begin{tabular}{lccccccccccc}
\toprule
Model & Bemba & Fijian & Faroese & Venetian & Tuvan & Sardinian & Banking & FewEvent & CLINC & CLINCD & FP \\
\midrule \midrule
4o-mini & 100 & 150 & 150 & 100 & 64 & 64 & 1000 & 2000 & 2000 & 2000 & 500 \\
\hline
4o      & 100 & 150 & 100 & 200 & 100 & 100 & 1000 & 500  & 500  & 1000 & 100 \\
\bottomrule
\end{tabular}
}
\caption{Optimal demonstration counts for Naive-SemiICL per dataset under 4o-mini and 4o models.}
\label{tab:best_shot_all}
\end{table*}

\subsection{Baseline Details} \label{appendix:baseline_det}
We observe that few-shot baselines (Section \ref{sec:experiments}) does not necessarily scale with more demonstrations. Thus, we report the best performing $k$-shot baseline where $k \leq 16$, which we report in Table \ref{tab:best_shot}.

We experiment with different pseudo-demonstration sizes $k_u$ for Naive-SemiICL: $k_u \in \{32, 64, 100, 150, 200\}$ on classification tasks, and $k_u \in \{32, 64, 100, 150, 200\}$ on translation and reasoning tasks.

For MoT \cite{li-qiu-2023-mot}, we follow a recommended configuration of $5$ clusters. For retrieval, we employ the same text embeddings, \texttt{text-embedding-3-large} as Naive-SemiICL, and the same confidence threshold set at the 90th percentile. Since MoT needs to query the LLM $k$ times to select the most relevant examples, it is not suitable for classification and translation tasks that might utilize many examples, we only compare MoT to Naive-SemiICL on reasoning tasks.

\subsection{Applying Confidence Scores} \label{appendix:confidence_det}
On all tasks, we sample from the LLMs 10 times to compute the Self-Consistency score. Self-Consistency is unsuitable for translation tasks due to the computational challenges of assessing equivalence between translations. Instead, we introduce Back-Translation, which evaluates translation quality by translating the output back to the original language. The confidence score is then derived using the cosine similarity (on embeddings) between the back-translation and the original input. A detailed description of Back-Translation is provided in Appendix \ref{appendix:back-translation}. \\

\subsection{Pseudo-Demonstration Sampling} \label{appendix:even_sampling}
On classification tasks, BigBenchHard tasks and GPQA, we sample diversely by evenly sample predicted labels. For translation tasks and LiveBench Math, we utilize OpenAI's \texttt{text-embedding-3-large} to generate embeddings for each example input. Then we cluster the embeddings into clusters and evenly sample the most representative (the one closest to the cluster centroid) instances from each cluster.

\subsection{Back-Translation} \label{appendix:back-translation}
Suppose an LLM has translated a source language input $s$ into a target language output $t$. We then use the same LLM to translate $t$ back to the original language
\begin{equation*}
    \hat{s} = \mathrm{LM}(t, \rho_b),
\end{equation*}
where $\rho_b$ is a prompt that induces the back-translation. Then, the Back-Translation Confidence is the cosine similarity between the original input $s$ and the back-translation $\hat{s}$
\begin{equation*}
    c = \mathrm{sim}_\mathrm{cos}(\phi(\hat{s}), \phi(s)),
\end{equation*}
where $\phi$ is an embedding function.

\subsection{Hyperparameters} \label{appendix:iterpsd_hyper}
\paragraph{Confidence Threshold.} We find that setting the confidence threshold $\lambda$ at the 90th percentile of all generate pseudo-demonstrations $\mathcal{D}_\mathrm{PSD}$ allowed Naive-SemiICL to achieve competitive performance. We assume this threshold in this work unless stated otherwise.

\paragraph{IterPSD} takes 3 hyperparameters: chunk size $K$, which controls how many data to annotate in each iteration, $\epsilon$, which controls the proportion of random sample in each chunk, $\kappa$, the maximum amount of examples allowed in the demonstration while generating pseudo-demonstrations. We experiment with $K \in \{100, 500, 1000\}$, $\epsilon \in \{ 0.5, 0.8, 1.0 \}$, $\kappa \in \{ 300, 500, 1000\}$. We find that $\epsilon = 0.8$, $K = 500$, and $\kappa = 1000$ yielded the best results on all tasks, except on FP, where $K = 100$ and $\kappa = 300$ yielded the best result.

\section{Effects of Different Confidence Methods} \label{appendix:confidence}
In this section, we examine the performance Naive-SemiICL paired with different confidence methods, which we compile as Table \ref{tab:confidence}. We observe that classification and translation tasks each have a dominant confidence measure. For classification tasks, Self-Consistency emerges as the most effective confidence method. It surpasses the Verbalized Confidence method on 4 out of 5 datasets across both models. Verbalized Confidence is the leading measure for translation tasks, consistently achieving the highest performance across all languages. For reasoning tasks, no single method clearly dominates. Under GPT-4o-mini, Verbalized Confidence yields the best average performance, while under GPT-4o, Entropy slightly outperforms Self-Consistency, securing the top position by a narrow margin.

Overall, Self-Consistency improves classification and reasoning tasks, but its effect varies across translation tasks and is not applicable to all tasks. Entropy is sometimes useful in reasoning tasks, but fall short on translation tasks. Verbalized inference remains a strong and economical baseline across all tasks but is generally outperformed by Self-Consistency on classification tasks.

\begin{table*}[!th]
\caption{The prompt template we use for classification, translation, and reasoning tasks, respectively.}
\label{tab:prompts}
\small
\begin{tabularx}{\textwidth}{>{\centering\arraybackslash}m{3cm}m{12cm}}
\toprule
\textbf{Types} & \textbf{Prompts} \\ 
\midrule \midrule
\quad & \quad \\
Classification &
You are a helpful assistant who is capable of performing a classification task (mapping an Input to a Label) with the following possible labels:

\{A LIST OF POSSIBLE LABELS\}

\_\_\_

Here are zero or more Input and Label pairs sampled from the classification task.

\quad

\{DEMONSTRATIONS\}

\quad

\_\_\_

Now, Label the following Input among the following

Input: \{INPUT\} \\
\quad & \quad \\
\hdashline
\quad & \quad \\
Translation &       You are an expert translator. I am going to give you zero or more example pairs of text snippets where the
first is in the source language and the second is a translation of the first snippet into the target language. The sentences will be written in the following format:

<source language>: <first sentence>

<target language>: <translated first sentence>

\_\_\_

\quad

\{DEMONSTRATIONS\}

\quad

\_\_\_

Now, Translate the following \$source text into \$target. Also give the Confidence of your given Answer in the following format: 

**Confidence**: <a confidence score between 0 and 1>:

\quad

English: \{INPUT SENTENCE\}

\{TARGET LANGUAGE\}:  
\\
\quad & \quad \\
\hdashline
\quad & \quad \\
Reasoning & 
First, I am going to give you a series of Questions that are like the one you will be solving.

\_\_\_

\quad

\{DEMONSTRATIONS\}

\quad

\_\_\_

Now, Answer the following Question. Think step by step.

Question: \{QUESTION\}

Also give the Confidence of your given Answer in the following format: 

**Confidence**: <a confidence score between 0 and 1> \\
\quad & \quad \\
\hline
\end{tabularx}
\end{table*}

\end{document}